\documentclass[12pt]{article}

\pdfoutput=1
\usepackage{adjustbox}
\usepackage{algorithm}
\usepackage{algpseudocode}
\usepackage{amsfonts}
\usepackage{amsmath}
\usepackage{amssymb}
\usepackage{amsthm}
\usepackage{bm}
\usepackage{boldline}
\usepackage{booktabs}
\usepackage{cellspace}
\usepackage{caption}
\usepackage{ctable}
\usepackage{color}
\usepackage[T1]{fontenc}
\usepackage[margin=1.375in]{geometry}
\usepackage{graphicx}
\usepackage{hhline}
\usepackage[hidelinks]{hyperref}
\usepackage[utf8]{inputenc}
\usepackage{multirow}
\usepackage[authoryear]{natbib}
\usepackage{numprint}
\usepackage{subcaption}
\usepackage{scalerel}
\usepackage{url}

\usepackage{fullpage}
\usepackage{fancyhdr}
\fancypagestyle{firstpage}{
\chead{To appear in Neural Computation, 2020}
}

\title{\LARGE Resonator networks for factoring distributed representations of data structures}
\author{
\begin{tabular}[t]{c@{\extracolsep{3ex}}c} 
E. Paxon Frady$^{1,5}$ & Spencer J. Kent$^{1,2}$\\
\rule{0pt}{4.5ex}
Bruno A. Olshausen$^{1,3,4}$ \hspace{0.25ex} & Friedrich T. Sommer$^{1,3,5}$   \\
\end{tabular}
}
\date{\vspace{0.2cm}\small{
$^1$Redwood Center for  Theoretical Neuroscience \\
$^2$Electrical Engineering and Computer Science \\
$^3$Helen Wills Neuroscience Institute \\
$^4$School of Optometry \\
      University of California, Berkeley \\
      Berkeley, CA 94702 \\
\vspace{0.25cm}
    $^5$Intel Laboratories, Neuromorphic Computing Lab\\
    San Francisco, CA 94111
\vspace{0.25cm}}}
\begin{document}

\maketitle
\thispagestyle{firstpage}

\begin{abstract}
The ability to encode and manipulate data structures with distributed neural representations could qualitatively enhance the capabilities of traditional neural networks by supporting rule-based symbolic reasoning, a central property of cognition. 
Here we show how this may be accomplished within the framework of Vector Symbolic Architectures (VSA) \citep{plate1991holographic,gayler1998multiplicative,kanerva1996binary}, whereby data structures are encoded by combining high-dimensional vectors with operations that together form an algebra on the space of distributed representations. 
In particular, we propose an efficient solution to 
a hard combinatorial search problem that arises when decoding elements of a VSA 
data structure: the factorization of products of multiple code vectors. 
Our proposed algorithm, called a
resonator network, is a new type of recurrent neural network that interleaves VSA multiplication operations and 
pattern completion. 
We show in two examples -- parsing of a tree-like data structure and parsing of a visual scene -- how the factorization problem arises and how the resonator network can solve it.  More broadly,  resonator networks open the possibility to apply VSAs to myriad artificial intelligence problems in real-world domains.
A companion paper \citep{kent2020resonator} presents a rigorous analysis and evaluation of the performance of resonator networks, showing it out-performs alternative approaches. 
\end{abstract}

\newpage
\section{Introduction}
\label{sec:intro}
Cognition requires making use of learned knowledge in contexts never before encountered, a facility that requires information to be represented in terms of components that may be flexibly recombined.
A longstanding goal for neuroscience and psychology
has been to understand how such capacities are expressed by neural networks in the brain. 
Early artificial intelligence researchers 
developed frameworks of symbol-manipulation to emulate cognition, but they were implemented with local data representations (where the meaning of a bit is tied to its location) that are brittle and non-adaptive \citep{kanerva1997fully}.  
Connectionism, a movement started in psychology \citep{mcclelland1986parallel},
based itself on the premise that internal representations of knowledge must be
highly distributed and be able to adapt to the statistics of the data, so as to learn by example.  Along the way, however, connectionism also gave up many of the rich capabilities offered by symbolic computation \citep{jackendoff2002}.
In recent years, it has become clear that 
a unification of the ideas behind each approach -- distributed representation, adaptivity, 
and symbolic manipulation -- will be required for reproducing the brain's ability
to learn from few examples, to deal with novel situations, or to change behaviors when driven by internal information processing rather than purely by external events. \citep{plate2003holographic, gayler2004vector, kanerva2009hyperdimensional, lake2017building}.

Digital computers owe their power and ubiquity to the abstraction of \textit{data structures}, which support decomposing information into parts,
referencing each part individually, and composing these parts with other data structures.  Examples include trees, records with fields, or linked lists.
Connectionist theories have long been criticized because it is hard to imagine how compound, hierarchical data structures could be represented and manipulated by neural networks \citep{hinton1990mapping}. 
Cognitive scientists have argued that, at the very least, cognitive data structures should support three patterns of combination, which are familiar to any computer programmer \citep{fodor1988connectionism}. 
1) Key-value pairs: A key or variable is a placeholder for information to which a value can be assigned in a particular instance. 
This association, \emph{variable binding}, generates what is called the systematicity of cognition \citep{fodor1975language, plate2003holographic}. 
2) Sequential structures: 
A \emph{sequence} is an ordered pattern of organization and computation required by many reasoning tasks. 
3) Hierarchy: the notion that some aspects of knowledge can be decomposed recursively into a set of successively more fundamental parts. Variable binding, sequence, and hierarchy are critical structures of cognition, and a comprehensive theory of intelligence must take these into account.

A family of models called Vector Symbolic Architectures (VSAs) encode these structures into distributed representations, providing a framework that can reconcile the symbolic and connectionist perspectives \citep{plate2003holographic, gayler2004vector, kanerva2009hyperdimensional}.
Building on the concept of reduced representations \citep{hinton1990mapping}, VSAs allow one to express data structures holographically in a vector space of high but fixed dimensionality.
The atoms of representation are random high-dimensional vectors, and data structures built from these atoms are vectors with the same dimension. 
Three operations are used to form and manipulate data structures -- addition, multiplication, and permutation -- which together form an algebra over the space of high-dimensional vectors.  These operations enable building representations of sets, ordered lists (sequences), n-tuples, trees, key-value bindings, and records containing role-filler relationships which can be composed into hierarchies, as described in \citet{plate1995holographic, kanerva1996binary, kanerva1997fully, joshi2016language, Frady2018} and below.

In order to read out or access the components of a VSA-encoded data structure, the high-dimensional vector representing it must be decomposed into the primitives or atomic vectors from which it is built.
This is the problem of \textit{decoding}.
For example, if the primitives are combined by addition only, the distributed representation can be decoded by a nearest-neighbor lookup or an autoassociative memory. 
However, hierarchical or compound data structures, such as a multi-level tree or an object with multiple attributes bound together, are built from combinations of addition, multiplication and permutation operations on the primitives.
In this case, decoding via a simple nearest-neighbor lookup would require storing every possible combination of the primitives -- e.g., all possible paths in a tree, or all the possible attribute combinations -- essentially amounting to a combinatoric search problem. Past applications of VSAs have largely sidestepped this problem by limiting the depth of the data structures or using a brute force approach to consider all possible combinations when necessary  \citep{plate2000analogy, cox2011toward}.
As a result, the application of VSAs to real-world problems has been rather limited, since up to now there has not been a solution for efficiently accessing elements of such compound data structures containing a product of multiple components.

The solution to this dilemma is to \emph{factorize} the high-dimensional vector representing a compound data structure into the primitives from which it is composed.  That is, given a high-dimensional vector formed from an element-wise product of two or more vectors, we must find its factors.  This way, a nearest-neighbor lookup need only search over the alternatives for each factor individually rather than all possible combinations.
Obviously though, factorization poses a difficult computational problem in its own right.  

Here, we propose an efficient algorithm for factorizing high-dimensional vectors that may be interpreted as a type of recurrent neural network, which we call a \emph{resonator network}.
The resonator network relies on the VSA principle of \emph{superposition} to search through the combinatoric solution space without directly enumerating all possible factorizations.
Given a high-dimensional vector as input, the network iteratively searches through many potential factorizations in parallel until a set of factors is found that agrees with the input. Solutions emerge as stable fixed-points in the network dynamics. 

In this paper, part one of a two-part series, we first briefly introduce the VSA framework and the problem of factoring high-dimensional VSA representations. We then show via two examples -- searching a binary tree and querying the contents of a visual scene -- how VSAs may be used to build distributed representations of compound data structures, and how resonator networks are used to decompose these data structures and solve the problem. Part two of this series \citep{kent2020resonator} provides rigorous mathematical and simulation analysis of resonator networks, and compares its performance with alternative approaches for solving high-dimensional vector factorization problems. 

\section{VSA preliminaries}
\label{sec:represent}

All entities in a VSA are represented as high-dimensional vectors in the same space, with vector dimension $N$ typically in the range of $1{,}000-10{,}000$. In this paper, 
we focus on the VSA framework called \emph{Multiply-Add-Permute} \citep{gayler1998multiplicative, gayler2004vector}. The atomic primitives are `bipolar' vectors whose components are $\pm 1$, chosen randomly. These vectors are used as symbols to represent concepts. The set of atomic vectors representing specific items are stored in a codebook, which is a matrix of dimension $N \times D$, where $D$ is the number of atoms.

The use of high-dimensional vectors is an important aspect of the VSA framework, as it relies on the concentration of measure phenomenon \citep{ledoux2001concentration} that independently chosen random vectors are very close to orthogonal, a property we refer to as \emph{quasi-orthogonality}. This property allows vectors to act symbolically, as the similarity (inner product) between two different atomic vectors is small compared to their self-similarity (L2 norm). Furthermore, a much larger set of quasi-orthogonal vectors exist than orthogonal vectors, which may be exploited for combinatoric search.

Data structures are composed and computations are carried out via an algebra consisting of three vector operations: addition, multiplication and permutation.
The elements of a data structure are then read out (decoded) using the conventional vector dot product as a similarity measure to compare to items stored in the codebook.  The VSA operations of addition, multiplication, and permutation act to manipulate the vector symbols in ways that preserve or destroy their similarity.

\vspace{0.1in}
Formally, the VSA operations are defined as follows:  
\begin{description}

\item[Dot product ($\cdot$)] is the conventional vector inner product, $\mathbf{x}\cdot\mathbf{s}=\sum_i x_i\,s_i$, which is used to measure the `similarity' between vectors. This is used to decode the result of a VSA computation by comparing the vector to the set of vectors in the codebook:
\begin{equation*}
    \mathbf{a} = \mathbf{X}^\top \mathbf{s}
\end{equation*}
Here, $\mathbf{X}$ is the codebook of atomic vectors, and $\mathbf{s}$ is a high-dimensional vector resulting from a VSA computation. 
The result of a VSA computation can be a single symbol indicated by the largest component of $\mathbf{a}$. Alternatively, the coefficients $\mathbf{a}$ can be considered as a weighted sum, where each entry indicates a confidence level, probability, or intensity value. 

\item[Addition ($+$)] is used to `superpose' items together, like forming a set. It is defined by regular vector addition, the element-wise sum: 
\[
\mathbf{s} = \mathbf{x} + \mathbf{y},
\]
or $s_i = x_i+y_i$. Depending on the circumstances, the sum may be kept as is, or subsequently thresholded so that each $s_i$ is $\pm 1$. In either case, the addition operation results in a vector that is similar to each of its superposed components -- one can determine the members of the sum by similarity to the atomic vectors. 
Superposition is possible because of the quasi-orthogonal property. However, superposition produces a small amount of \emph{crosstalk noise}, which increases with the number of items in the sum and is diminished with large vector dimensionality (see \citet{Frady2018} for detailed characterization of superposition).

\item[Multiplication ($\odot$)] is used to `bind' items together to form a conjunction, such as in assigning a value to a variable. It is defined by the Hadamard product between vectors, i.e. the element-wise multiplication of vector components:
\[
\mathbf{s} = \mathbf{x} \odot \mathbf{y},
\]
or $s_i=x_i\, y_i$. This multiplication operation is invertible, i.e. $\mathbf{y} = \mathbf{s} \odot \mathbf{x}$, and it distributes over addition, $\mathbf{x} \odot \mathbf{y} + \mathbf{x} \odot \mathbf{z} = \mathbf{x} \odot (\mathbf{y} + \mathbf{z})$. 
Note that in the MAP VSA, the bipolar primitive vectors are their own self-inverses.
In contrast to addition, multiplication generates a vector that is dissimilar to each of its inputs \citep{kanerva2009hyperdimensional}. 

\item[Permutation ($\rho(\cdot)$)] is used to `protect' or `order' items. Permutation operates on a single input vector.
In principle, it can be any random permutation, but is typically a simple cyclic shift:
\[
\mathbf{s} = \rho(\mathbf{x})
\]
or $s_i = x_{(i-1) \% N}$.
Permutation distributes over both addition, $\rho(\mathbf{x}) + \rho(\mathbf{y}) = \rho(\mathbf{x} + \mathbf{y})$,
and multiplication, $\rho(\mathbf{x}) \odot \rho(\mathbf{y}) = \rho(\mathbf{x} \odot \mathbf{y})$, and its function is complementary to addition and multiplication. 
Permutations are used to protect the components of a data structure built with these other operations, based on the fact that permutation and binding are non-commutative, $\mathbf{x} \odot \rho(\mathbf{y}) \ne \mathbf{y} \odot \rho(\mathbf{x})$. 
In essence, permutation rotates vectors into dimensions of the space that are almost orthogonal to the dimensions used by the original vectors.
Information is thus protected when combined with other items, because vector components will not appear similar to or interfere with those other items. Permutations can also be used to index sequences \citep{Frady2018}, or levels in a hierarchy, by successive application of the permutation operation. For example to represent the sequence $\mathbf{x}_0, \mathbf{x}_1, \mathbf{x}_2$ in a vector $\mathbf{s} = \mathbf{x}_0 + \rho(\mathbf{x}_1) + \rho^2(\mathbf{x}_2)$, with $\rho^2(\mathbf{x}) = \rho( \rho(\mathbf{x}))$.
\end{description}

VSAs combine these operations to form data structures and to compute with them. 
The combination of atomic vectors into composite data structures is rather straightforward. But, as we shall see, querying composite data structures often results in the problem of decoding terms composed of two (or perhaps many more) atomic vectors that are multiplied together. In order to decode such composite vectors, one must search through many combinations of atoms. In general, this is a hard combinatorial search problem, which typically requires directly testing every combination of factors. The resonator network can efficiently solve these problems without needing to directly test every combination of factors.

\section{Factorization via search in superposition}
\label{sec:resonator}

In general, the factorization problems that arise in VSAs may involve two or more factors, but
let us assume we are given a composite vector $\mathbf{s}$, formed as a product of three vectors:
\begin{equation}
    \mathbf{s} = \mathbf{x}_{i^*} \odot \mathbf{y}_{j^*} \odot \mathbf{z}_{k^*}
\label{eqn:basic_res_problem}
\end{equation}
where the vectors $\mathbf{x}_{i^*}$, $\mathbf{y}_{j^*}$, and $\mathbf{z}_{k^*}$ are drawn from codebooks $\mathbf{X}=\{ \mathbf{x}_{1}, ..., \mathbf{x}_{D}\}$, $\mathbf{Y}=\{ \mathbf{y}_{1}, ..., \mathbf{y}_{D}\}$ and $\mathbf{Z}=\{ \mathbf{z}_{1}, ..., \mathbf{z}_{D}\}$. 
Given $\mathbf{s}$ and the codebooks $\mathbf{X}$, $\mathbf{Y}$, and $\mathbf{Z}$, the task is to find $\mathbf{x}_{i^*}$, $\mathbf{y}_{j^*}$, and $\mathbf{z}_{k^*}$. 

The resonator network is an iterative approach to solve this problem without exhaustively searching through each possible combination of the factors. A key motivating idea behind resonator networks is the VSA principle of superposition. In VSAs, multiple symbols can be expressed simultaneously in a single high-dimensional vector via vector addition. Randomized atomic vectors are highly likely to be close to orthogonal in high-dimensional space, meaning that they can be superposed without much interference. However, there is some crosstalk noise between the superposed symbols, and  ``clean-up memory'' (such as a Hopfield Network) is thus utilized to reduce the crosstalk noise.

A resonator network combines the strategy of superposition and clean-up memory to efficiently search over the combinatorially large space of possible factorizations. 
The vectors $\mathbf{\hat{x}}$, $\mathbf{\hat{y}}$, and $\mathbf{\hat{z}}$ represent the current estimate for each factor. 
These vectors can be initialized to the superposition of all possible factors, e.g. $\mathbf{\hat{x}}(0) = \sum_i^D \mathbf{x}_i$, $\mathbf{\hat{y}}(0) = \sum_j^D \mathbf{y}_j$, etc. 
A particular factor can then be inferred from $\mathbf{s}$ based on the estimates for the other two, e.g. $\mathbf{\hat{z}}(1) = \mathbf{s} \odot \mathbf{\hat{x}}(0) \odot \mathbf{\hat{y}}(0)$. 
Since binding distributes over addition, the product $\mathbf{\hat{x}}(0) \odot \mathbf{\hat{y}}(0)$ expresses every combination of factors in superposition, because $\mathbf{\hat{x}}(0) \odot \mathbf{\hat{y}}(0) = \sum_i^D \sum_j^D \mathbf{x}_i \odot \mathbf{y}_j$.
For instance, if $D=100$, then this initial guess represents $D^2 = 10{,}000$ combinations in superposition. Thus, many potential combinations of the pair of factors may be considered at once when inferring the third factor. 

The inference process, however, is noisy if many guesses are tested simultaneously. 
This noise results from crosstalk of many quasi-orthogonal vectors, and can be reduced through a clean-up memory. This is built from the codebooks, which contain all the vectors that are possible factors of the input $\mathbf{s}$.
Each clean-up memory projects the initial noisy estimate onto the span of the codebook. This computes a measure of confidence for whether each element in the codebook is a factor.

The result of the inference and clean-up leads to a new estimate for each factor. The new estimate is formed by a sum of dictionary items weighted by the confidence levels. This produces a better guess for each one of the factors. The inference can then be repeated with  better guesses, which reduces crosstalk noise even further.
By iteratively applying this procedure, the inference and clean-up stages cooperate to 
successively reduce crosstalk noise until the solution is found.
 \begin{figure}
    \centering
    \includegraphics[width=0.6\textwidth]{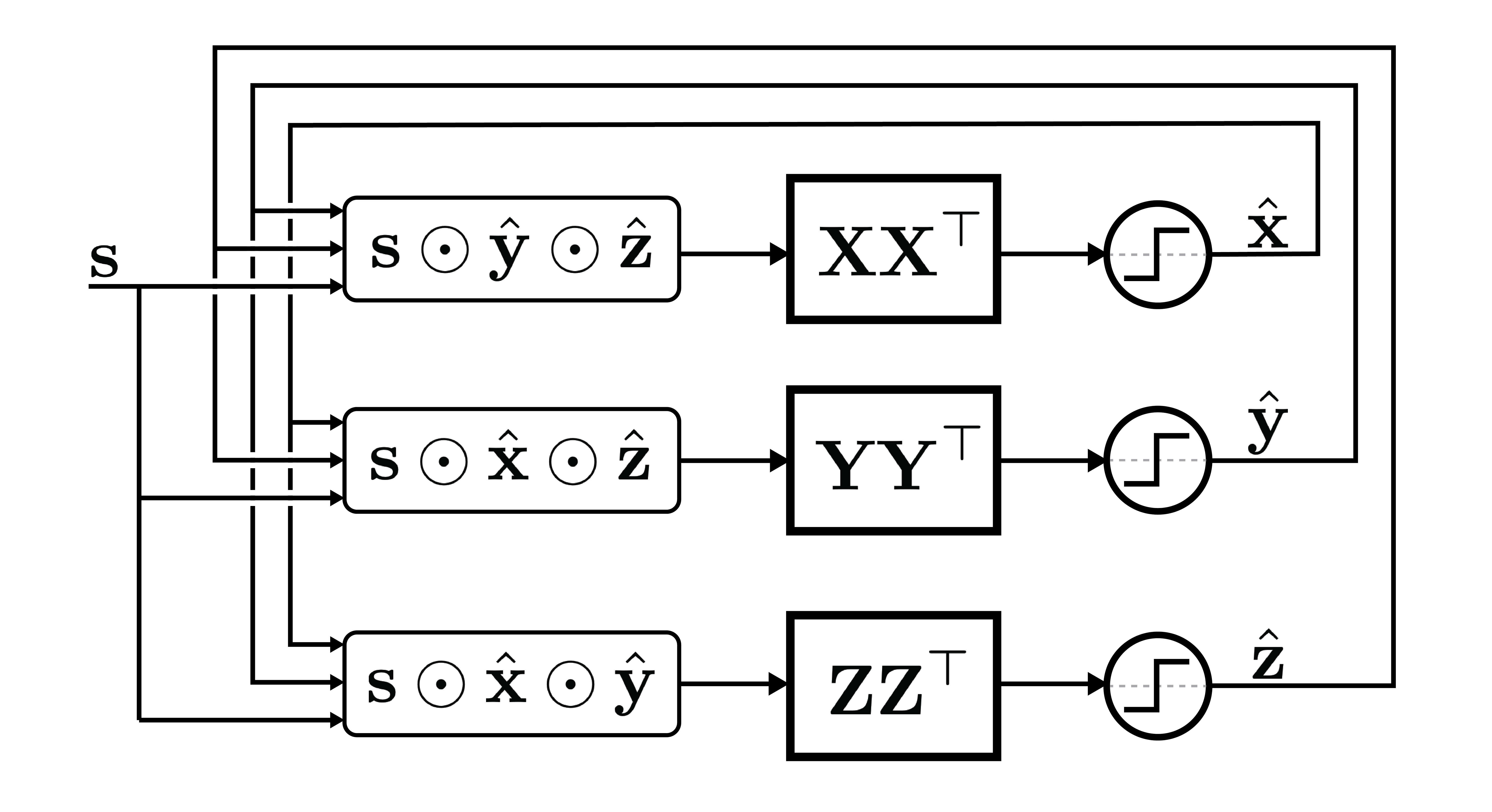}
    \caption{A resonator network with three factors}
    \label{fig:res_net_diagram}
\end{figure}

The procedure described above, for all three factors, is specified by the following set of equations (Fig. \ref{fig:res_net_diagram}):
\begin{align}
    \begin{split}
        \mathbf{\hat{x}}(t+1) &= g(\mathbf{X} \mathbf{X}^\top (\mathbf{s} \odot \mathbf{\hat{y}}(t) \odot \mathbf{\hat{z}}(t))) \\
        \mathbf{\hat{y}}(t+1) &= g(\mathbf{Y} \mathbf{Y}^\top (\mathbf{s} \odot \mathbf{\hat{x}}(t) \odot \mathbf{\hat{z}}(t))) \\
        \mathbf{\hat{z}}(t+1) &= g(\mathbf{Z} \mathbf{Z}^\top (\mathbf{s} \odot \mathbf{\hat{x}}(t) \odot \mathbf{\hat{y}}(t))) \\
    \end{split}
    \label{eq:three_fac_dynamics}
\end{align}
where the function $g$ prevents runaway positive feedback by thresholding the elements of each vector to $\pm1$. 

If we examine the clean-up memory for $\mathbf{\hat{x}}$, which contains a matrix multiplication with  $\mathbf{X} \mathbf{X}^\top$ and thresholding function $g$, then we see this operation is nearly identical to a Hopfield network with outer-product Hebbian learning \citep{Hopfield1982}. Except here, rather than directly feeding back into itself, the result of the clean-up is sent to other parts of the network.

The set of equations in (\ref{eq:three_fac_dynamics})  defines a nonlinear dynamical system that has interesting empirical and theoretical properties, which we thoroughly examine through simulation experiments in part two of this series \citep{kent2020resonator}. Empirically, the system bounces around in state space until the correct solution appears to ``resonate'' with the network dynamics, popping out as if in a moment of insight.
We find that while there is no Lyapunov function governing these dynamics and no guarantee for convergence, the resonator network empirically converges to the correct solution with high probability, as long as the number of product combinations to be searched is within the network's \emph{operational capacity}. We show that the operational capacity is given by a quadratic function of $N$. 
Compared to numerous alternative optimization methods that we considered, this capacity for resonator networks is higher by almost two orders of magnitude.   

\section{Decoding data structures with resonator networks}
\label{sec:decoding}

We now turn to two examples that illustrate how VSA operations can be combined to build distributed representations of data structures, how the factorization problem arises when parsing these representations, and how resonator networks can be designed to solve this problem.

\subsection{Searching a tree data structure}
\label{sec:tree_encoding}

\begin{figure}[t]
    \centering
    \includegraphics[width=0.9\textwidth]{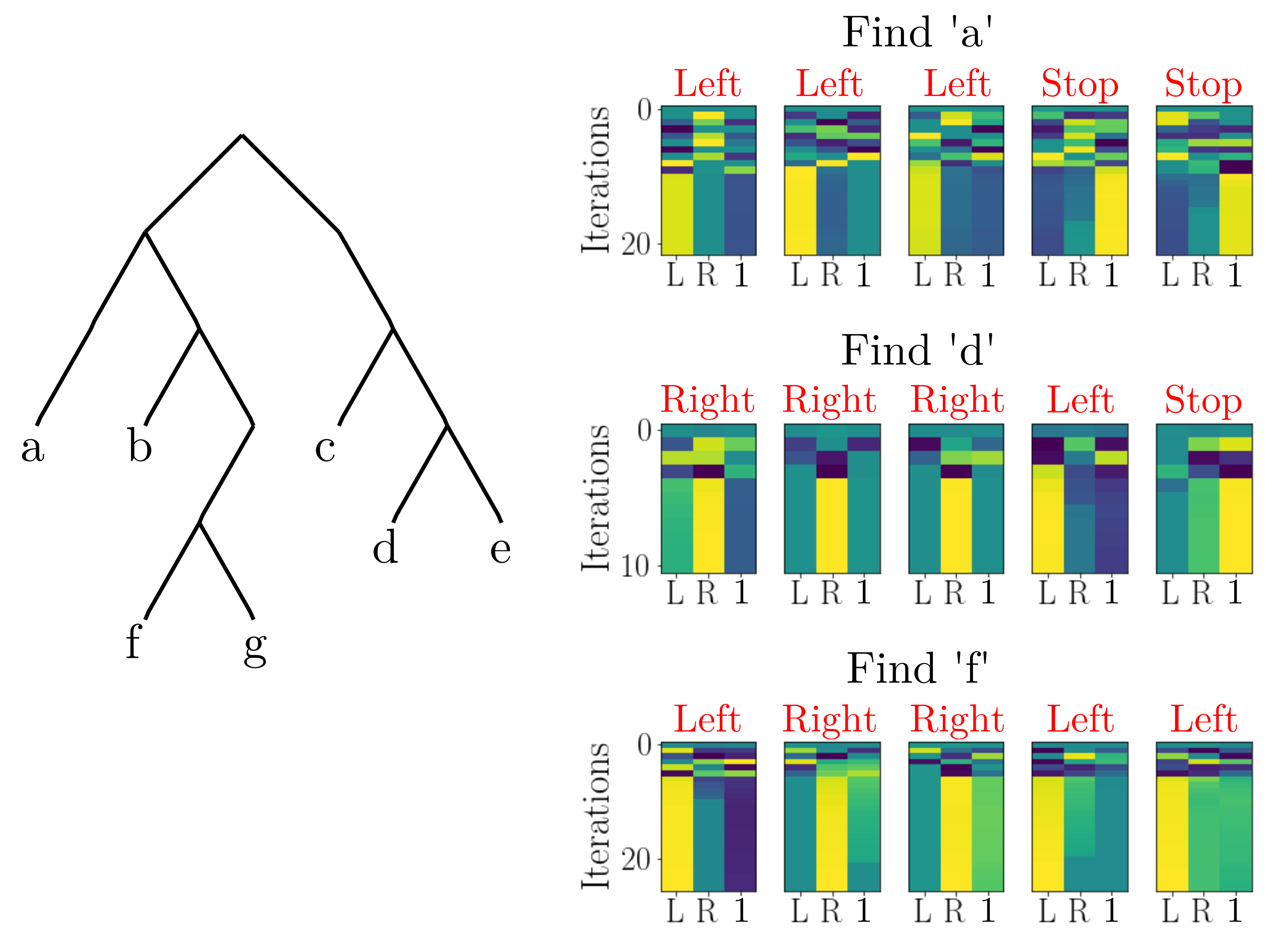}
    \caption{Tree search with a resonator network. The query of the vector $\mathbf{tree}$ produces an encoding of position which the resonator network can factor. The colored plots indicate the time evolution of $\hat{\mathbf{x}}^{(0)} \ldots \, \hat{\mathbf{x}}^{(4)}$ (from left to right), showing the cosine similarity of each estimate to each of the three possible vectors $\rho^{d}(\mathbf{left}), \rho^{d}(\mathbf{right}), \mathbf{1}$.
    Purple indicates low similarity, and yellow indicates high similarity. Initially the similarity changes significantly, until the three estimators find a coherent factorization and quickly converge. Red letters indicate the converged result for each $\hat{\mathbf{x}}^{(0)} \ldots \, \hat{\mathbf{x}}^{(4)}$.}
    \label{fig:res_tree}
\end{figure}

Consider the tree data structure depicted in Figure \ref{fig:res_tree}. We can form a distributed representation of this tree in a single high-dimensional vector by using all three VSA operations, namely superposition $+$, binding $\odot$, and permutation $\rho(\cdot)$. First, each leaf in the tree is assigned a random vector 
$\mathbf{a}, \mathbf{b}, \ldots, \mathbf{g} \in \{-1, +1\}^N$. 
We also assign random vectors $\mathbf{left}$ and $\mathbf{right}$ that are used to describe position in the tree. Moving from the root of the tree to a particular leaf involves a sequence of \texttt{left} and \texttt{right} turns. The order of these turns is represented by permutation $\rho(\cdot)$. The number of times permutation is applied
indicates depth within the tree: $\mathbf{left}$ is a \texttt{left} turn at depth $0$, $\rho(\mathbf{left})$ is a \texttt{left} turn at depth $1$, $\rho^2(\mathbf{left})$ is a \texttt{left} turn at depth $2$, and so on. A sequence of turns is represented by the \emph{binding} 
of these vectors, e.g., $\mathbf{left} \odot \rho(\mathbf{left}) \odot \rho^2(\mathbf{left})$ corresponds to three left turns. We can then attach to each leaf its position in the tree, again with binding, e.g., 
$\mathbf{a} \odot \mathbf{left} \odot \rho(\mathbf{left}) \odot \rho^2(\mathbf{left})$. Finally, the representation for the whole tree is collapsed into a single vector, $\mathbf{tree}$, via superposition:
\begin{align}
\begin{split}
\mathbf{tree} \,\,\, = \,\,\,\, & \,\mathbf{a} \odot \mathbf{left} \odot \rho (\mathbf{left}) \odot \rho^2 (\mathbf{left}) \\
+ \,&\, \mathbf{b} \odot \mathbf{left} \odot \rho (\mathbf{right}) \odot \rho^2 (\mathbf{left}) \\
+ \,&\, \mathbf{c} \odot \mathbf{right} \odot \rho (\mathbf{right}) \odot \rho^2 (\mathbf{left}) \\
+ \,&\, \mathbf{d} \odot \mathbf{right} \odot \rho (\mathbf{right}) \odot \rho^2 (\mathbf{right}) \odot \rho^3 (\mathbf{left})\\
+ \,&\, \mathbf{e} \odot \mathbf{right} \odot \rho (\mathbf{right}) \odot \rho^2 (\mathbf{right}) \odot \rho^3 (\mathbf{right})\\
+ \,&\, \mathbf{f} \odot \mathbf{left} \odot \rho (\mathbf{right}) \odot \rho^2 (\mathbf{right}) \odot \rho^3 (\mathbf{left}) \odot \rho^4 (\mathbf{left})\\
+ \,&\, \mathbf{g} \odot \mathbf{left} \odot \rho (\mathbf{right}) \odot \rho^2 (\mathbf{right}) \odot \rho^3 (\mathbf{left}) \odot \rho^4 (\mathbf{right})\\
\end{split}
\end{align}

The vector $\mathbf{tree}$ encodes the information so that we can flexibly query the data structure using VSA operations. For instance, we can find the identity of the leaf located at position \texttt{left}, \texttt{right}, \texttt{left} by ``unbinding'' the representation of this location from the vector representing the tree. 
Binding and unbinding are performed with the same operation since bipolar vectors are self-inverses. When we unbind the query location by Hadamard product, it will distribute through the superposition and cancel out with itself, leaving the atomic vector attached to that location ``exposed'':
\begin{equation}
    \mathbf{tree} \odot {\big(\mathbf{left} \odot \rho(\mathbf{right}) \odot \rho^2(\mathbf{left})\big)} = \mathbf{b} + noise
    \label{eq:tree_b_query}
\end{equation}
The noise term arises since the query distributes through the sum. The other terms combine with the query, but remain quasi-orthogonal to the vectors stored in the codebook, which keeps the other items in the tree ``hidden''. That is, the terms contained in $noise$ are dissimilar from each of the atoms stored in the codebook, and this appears as Gaussian noise when decoding \citep{Frady2018}.
The vector $\mathbf{b} + noise$ will have high similarity with atom $\mathbf{b}$ in the codebook, and will be successfully decoded by nearest neighbor or associative memory lookup among the atoms with high probability.

With this flexible encoding of the data structure, instead of asking for the label at a specific position, we can ask for the position of a specific label (essentially the problem of tree search). For instance, the query that exposes the position of label \texttt{c} is simply: 
\begin{equation}
    \mathbf{tree} \odot \mathbf{c} = \mathbf{right} \odot \rho(\mathbf{right}) \odot \rho^2(\mathbf{left}) + noise
    \label{eq:tree_c_query}
\end{equation}
This presents a new challenge, however, because we still need to decode the composite vector $\mathbf{right} \odot \rho(\mathbf{right}) \odot \rho^2(\mathbf{left}) + noise$ into the parts that describe a position in the tree. In previous applications of VSAs one would exhaustively enumerate all traversals of the tree and compute similarity to find the path. Instead, we can use a resonator network.

To set up the network for this problem, we first establish a maximum depth to search through -- the maximum depth determines the number of factors that need to be estimated. For the tree shown in Figure \ref{fig:res_tree}, we need five estimators, because this is the depth of the deepest leaves, \texttt{f} and \texttt{g}. 

Each factor estimate will determine whether to go \texttt{left}, \texttt{right} or to \texttt{stop}, for each level down the tree. To indicate \texttt{stop} a special vector is used, the identity vector $\mathbf{1}$ (a vector of all ones). By using the appropriate number of these identity vectors, each location in the tree can be thought of as a composition with the same depth (the maximum depth), even if the location is only partially down the tree. For instance, if we consider leaf \texttt{c} in the Figure \ref{fig:res_tree}, then its position $\mathbf{right} \odot \rho(\mathbf{right}) \odot \rho^2(\mathbf{left})$ is also
$\mathbf{right} \odot \rho(\mathbf{right}) \odot \rho^2(\mathbf{left}) \odot \mathbf{1} \odot \mathbf{1}$. This way, we can set up a resonator network for five factors and have it decode locations anywhere in the tree.

We denote each factor estimate as 
$\hat{\mathbf{x}}^{(0)}, \hat{\mathbf{x}}^{(1)}, \hat{\mathbf{x}}^{(2)}, \hat{\mathbf{x}}^{(3)}, \hat{\mathbf{x}}^{(4)}$ and the codebook matrices as 
$\mathbf{X}_0, \mathbf{X}_1, \mathbf{X}_2, \mathbf{X}_3, \mathbf{X}_4$. 
Each codebook matrix contains permuted versions of $\mathbf{left}$ and $\mathbf{right}$, and $\mathbf{1}$: $\mathbf{X}_d = \big[ \rho^{d}(\mathbf{left}), \rho^{d}(\mathbf{right}), \mathbf{1} \big]$
where $d$ indicates the depth in the tree. The network is constructed analogous to (\ref{eq:three_fac_dynamics}), but with five factor estimates running in parallel instead of three. For instance, the update equation for the first estimate is:
\begin{equation}
    \mathbf{\hat{x}}^{(0)}(t+1) = g \big(\mathbf{X}_0 \mathbf{X}_0 ^\top (\mathbf{s} \odot \mathbf{\hat{x}}^{(1)}(t)  \odot \mathbf{\hat{x}}^{(2)}(t) \odot \mathbf{\hat{x}}^{(3)}(t) \odot \mathbf{\hat{x}}^{(4)}(t)) \big)
\end{equation}

The process is demonstrated in Figure \ref{fig:res_tree}.
The input vector to be factorized, $\mathbf{s}$, is first formed from the tree data structure and the query. For instance to find the location of label \texttt{c}, $\mathbf{s} = \mathbf{tree} \odot \mathbf{c}$ is the input to the resonator network. Different leaves in the tree can be found by unbinding the leaf representation from the tree vector and using this result as the input. 

We visualize the network dynamics by displaying the similarity of each factor estimate $\hat{\mathbf{x}}^{(d)}(t)$ to the atoms stored in its corresponding codebook $\mathbf{X}_d$. The evolution of these similarity weights over time is shown as a heat map (Fig. \ref{fig:res_tree}, right). 
The heat maps show that the system initially jumps around chaotically, with the weighting of each estimate changing drastically each iteration. But then there is a quite sudden transition to a stable equilibrium, where each estimate converges nearly simultaneously, and at this point the output for each factor is essentially the codebook element with highest weight. 

\subsection{Visual scene analysis as a factorization problem}
\label{sec:encoding_viz_scenes}

\begin{figure}
    \centering
    \includegraphics[width=0.9\textwidth]{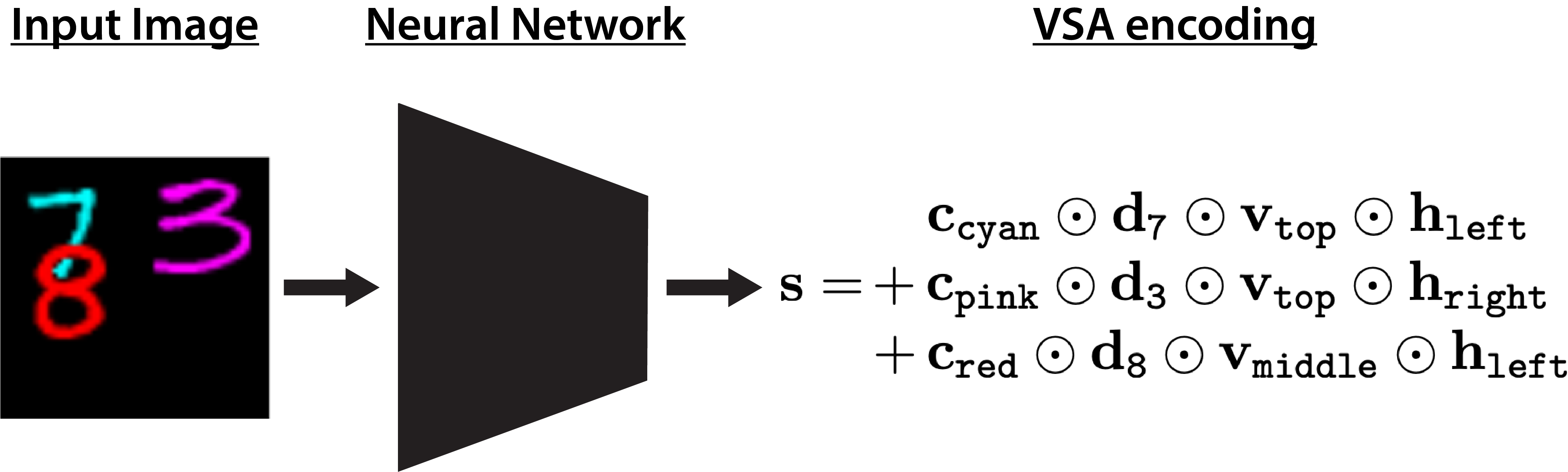}
    \caption{Generating a vector symbolic encoding of a visual scene}
    \label{fig:visual_scene_encoding_diagram}
\end{figure}

Next, we show how VSAs can encode the compositional structure of a visual scene, and how the resonator network can be used to decode the contents of the scene.  Consider the scene in Figure~\ref{fig:visual_scene_encoding_diagram} containing colored MNIST digits \citep{lecun1998mnist} in different positions. Position in the scene is indexed by vertical and horizontal coordinates, each quantized into three possible values,  (\texttt{top}, \texttt{middle}, \texttt{bottom}) and (\texttt{left}, \texttt{center}, \texttt{right}), respectively. Each digit can take on one of seven possible colors (\texttt{blue}, \texttt{green}, \texttt{cyan}, \texttt{red}, \texttt{pink}, 
\texttt{yellow}, \texttt{white}). The digits are labelled by their semantic class (\texttt{0}, \texttt{1}, \ldots, \texttt{9}), but the exact shape will differ, as the stimuli are sampled from the $50{,}000$ exemplars in the MNIST training set. 

Any given scene can have between one and three of these objects, which are allowed to partially occlude one another. We generate symbolic vectors $\mathbf{c}_{\texttt{blue}}, \mathbf{c}_{\texttt{green}}, \ldots, \mathbf{c}_{\texttt{white}}$ to encode color, 
$\mathbf{d}_{\texttt{0}}, \mathbf{d}_{\texttt{1}}, \ldots, \mathbf{d}_{\texttt{9}}$ to encode shape, $\mathbf{v}_{\texttt{top}}, \mathbf{v}_{\texttt{middle}}, \mathbf{v}_{\texttt{bottom}}$ to encode vertical position, and $\mathbf{h}_{\texttt{left}}, \mathbf{h}_{\texttt{center}}, \mathbf{h}_{\texttt{right}}$ to encode horizontal position, which are stored in respective codebooks, $\mathbf{C}, \mathbf{D}, \mathbf{V}, \mathbf{H}$.

The example scene (Fig. \ref{fig:visual_scene_encoding_diagram}) contains a \texttt{cyan} \texttt{7} at position \texttt{top}, \texttt{left}, a \texttt{pink} \texttt{3} at position 
\texttt{top}, \texttt{right}, and a \texttt{red} \texttt{8} at position \texttt{middle}, \texttt{left}. While this is a highly simplified type of visual scene, it illustrates the combinatorial challenge of representing and interpreting visual scenes. There are only $23$ distinct atomic
parameters ($10$ for digit identity, $7$ for color, $3$ each for vertical and horizontal position) and yet these combine to describe $10 \times 7 \times 3 \times 3 = 630$ individual objects, and
$630 + 630^2 + 630^3 = 250{,}444{,}530$ possible scenes with $1$, $2$, or $3$ objects. This number of combinations still does not include the variability among exemplars for each shape, of which there are $50,000$ in the MNIST dataset.

The VSA approach to represent a scene like this is to form the conjunction of each of the four factors with the binding operation, and superposing multiple objects together to form a single 
high-dimensional vector that constitutes a distributed representation of the entire scene. This encoding is depicted in Figure \ref{fig:visual_scene_encoding_diagram}, and like in the previous examples, the encoding provides a flexible data structure such that aspects of the scene can be individually queried. 
One attractive property of this representation is that its dimensionality does not grow with the number of objects in the scene, nor does it impose any 
particular ordering on the objects. 

To convert a new input image into a structured VSA representation, one challenge is to deal with the variability and correlations between the shapes of different hand-written digits. 
VSAs are designed for symbolic processing in neural networks. However, when dealing with sensor data streams one must solve the encoding problem, which is how to map the input data into the symbolic space \citep{rasanen2015generating, kleyko2018}.
We train a simple feed-forward  neural network with two fully-connected hidden layers to produce the desired VSA encoding of the scene. 
The feed-forward network was trained on a (uniformly) random sample of these scenes, with the MNIST digits chosen from an exclusive training set. A generative model creates the image of the scene from a random sample of factors for each object. From the chosen factors, the VSA representation of the scene is also generated through binding of VSA vectors for each factor and superposition for each object (Fig. \ref{fig:visual_scene_encoding_diagram}). Supervised learning via back-propagation is used to train the network to output the VSA representation of the entire scene from the image pixels as input. 

The resonator network can then be used to parse the output of the feed-forward network to identify each object and its properties.  
The vectors $\mathbf{\hat{c}}(t)$, $\mathbf{\hat{d}}(t)$, $\mathbf{\hat{h}}(t)$ and $\mathbf{\hat{v}}(t)$ denote the \emph{guesses} for each factor: color, digit, horizontal- and vertical-location, respectively. 
The scene can then be decoded by iterating through the resonator network:
 \begin{align}
\begin{split}
\mathbf{\hat{c}}(t+1) &= g \Big( \mathbf{C}\mathbf{C}^{\top} \Big( \mathbf{s} \odot \mathbf{\hat{d}}(t) \odot \mathbf{\hat{v}}(t) \odot \mathbf{\hat{h}}(t) \Big) \Big) \\
\mathbf{\hat{d}}(t+1) &= g \Big( \mathbf{D}\mathbf{D}^{\top} \Big( \mathbf{s} \odot \mathbf{\hat{c}}(t) \odot \mathbf{\hat{v}}(t) \odot \mathbf{\hat{h}}(t) \Big) \Big) \\
\mathbf{\hat{v}}(t+1) &= g \Big( \mathbf{V}\mathbf{V}^{\top} \Big( \mathbf{s} \odot \mathbf{\hat{d}}(t) \odot \mathbf{\hat{c}}(t) \odot \mathbf{\hat{h}}(t) \Big) \Big) \\
\mathbf{\hat{h}}(t+1) &= g \Big( \mathbf{H}\mathbf{H}^{\top} \Big( \mathbf{s} \odot \mathbf{\hat{d}}(t) \odot \mathbf{\hat{c}}(t) \odot \mathbf{\hat{v}}(t) \Big) \Big) \\
\end{split}
\label{eqn:resonator}
\end{align}

\begin{figure}[t]
    \centering
    \includegraphics[width=0.95\textwidth]{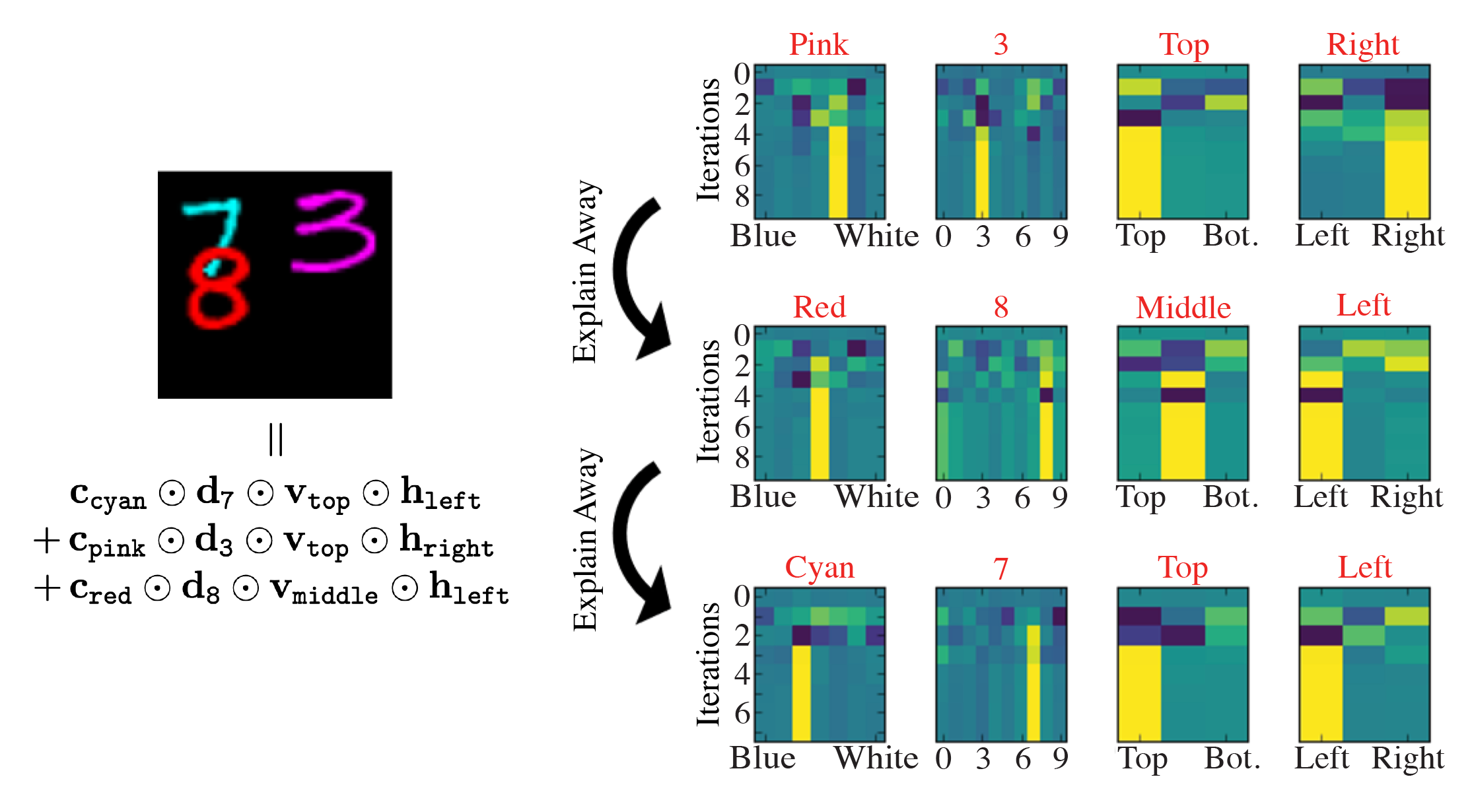}
    \caption{Scene vector $\mathbf{s}$ is fed into a resonator network which decodes each object in the scene. The model hones in on one object at a time, which is then explained away by subtracting the 
    resonator network's converged state from the scene vector. The network is reset and provided with this new input vector. It then converges to another solution, which describes a different object in the scene.}
    \label{fig:res_viz_scene}
\end{figure}

The encoding of visual scenes described superposes a composite vector for each object, each of which individually is a valid solution to the factorization of the scene. When we present the scene vector $\mathbf{s}$ to a resonator network, it automatically
hones in on a particular one of these composites, finding its factors. For instance, in Figure \ref{fig:res_viz_scene} the resonator network first identifies the pink three in the top right. Once the factorization has been found, this object is then ``explained away'' by subtracting it from $\mathbf{s}$.  What remains are the other composites, still in superposition. The resonator network is then reset (each resonator is reinitialized to the superposition of all possible codevectors) and presented with the new explained-away 
scene vector. It will then hone in on one of the remaining objects, in this case the red 8. This sequence may be repeated until all the objects have been decoded. This technique is similar to what is known as ``deflation'' in the context of 
tensor decomposition methods \citep{da2015iterative}.

After training on $100{,}000$ images, we used the network to produce symbolic 
vectors for a held-out test set of $10{,}000$ images. 
The vector dimensionality $N$ is a free parameter, which we chose to be $500$. If the exact ground truth vector is provided to a resonator network, 
it will infer the factors with $100\%$ accuracy provided $N$ is large enough, a fact we establish in part two of this series \citep{kent2020resonator}. 
For this small visual scene example it turns out $N=500$ more than suffices for the number of possible factorizations to be searched. Note that $N=500$ is less than the total number of combinations of all the factors, which is $630$.

\begin{figure}[t]
    \centering
    \includegraphics[width=0.8\textwidth]{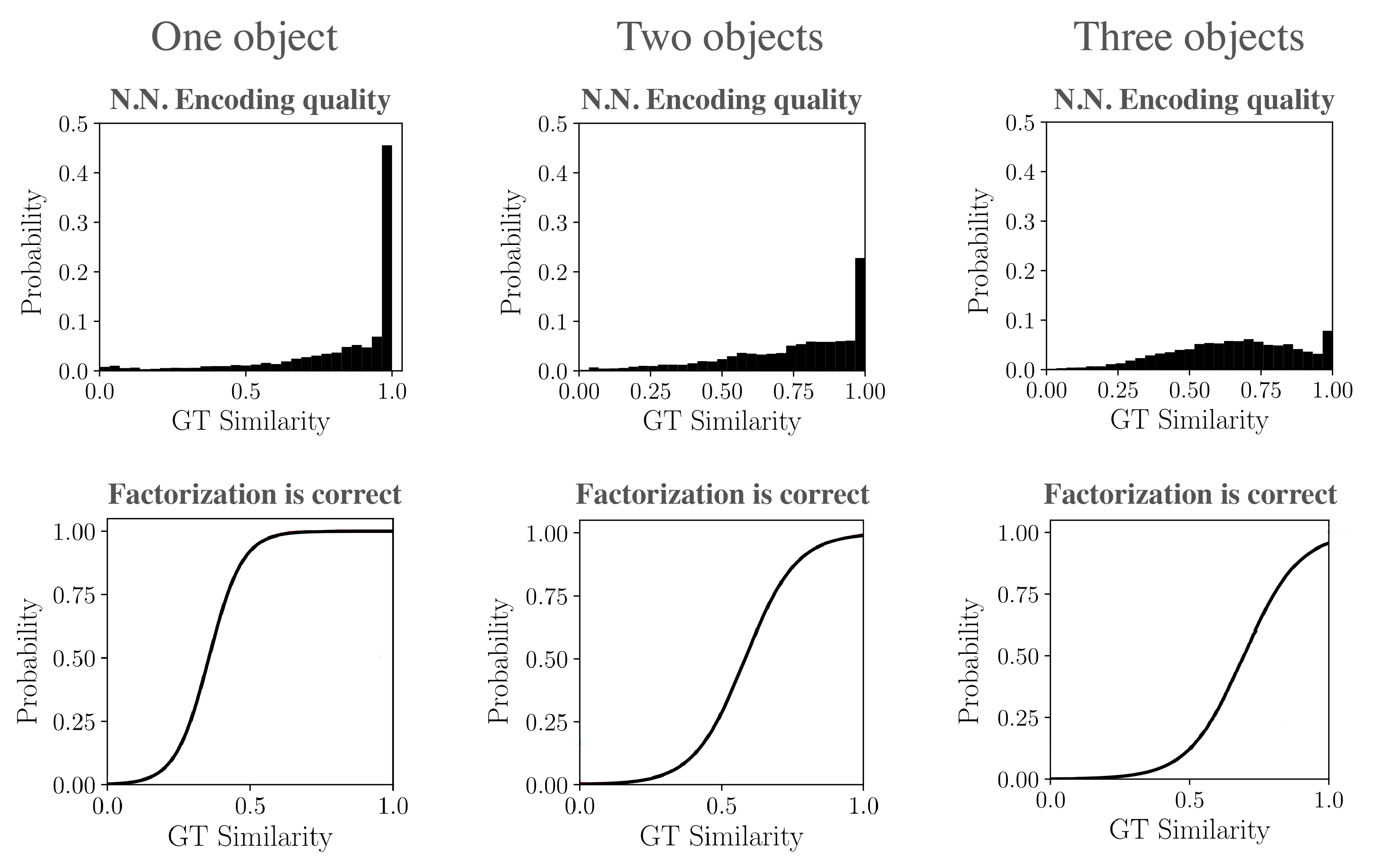}
    \caption{Resonator networks correct encoding errors. Visual scenes with one, two, and three objects are separated into separate columns. 
             Top row gives encoding quality, in terms of cosine similarity between the feed-forward network output and the ground-truth scene vector, across the test-set. 
             We define correct factorization as the case where the resonator network correctly infers all the factors of all objects. 
             The bottom row shows the empirical probability of a correct factorization as a function of similarity to the ground truth scene vector. Lines are logistic function fits to the data.}
    \label{fig:performance}
\end{figure}

The encoder network generates VSA scene vectors that are close to the ground-truth encoding, but there is some error. 
The error gets larger with more digits in the scene, perhaps partially due to occlusion of the digits. Figure \ref{fig:performance} shows that the resonator network can tolerate significant error in the scene vector produced by the feed-forward encoding network, correcting for ambiguity not resolved in the encoding step. 

\section{Discussion}

A major quest for modern artificial intelligence is to build computational models that combine the abilities of neural networks with the abilities of rule-based reasoning. Vector Symbolic Architectures, a family of connectionist models, enable the formation of distributed representations of data structures, structured computation on these representations, and has provided valuable conceptual insights for cognition and computation. However, so far, VSA models have not been able to solve challenging artificial intelligence problems in real-world domains
due to the combinatorial factorization problem that arises when processing complex, hierarchical data structures.  
Our contribution here has been to provide an efficient solution to the factorization problem, the resonator network, which we show in a companion paper \citep{kent2020resonator} vastly outperforms standard optimization methods. 

The two applications we showed here -- parsing a tree-like data structure and decomposing a visual scene -- are intended as illustrative examples to show how factorization of multi-term products arises in querying a VSA data structure, and they show how to design resonator networks to solve such problems.  Having a solution to the factorization problem now makes it possible to apply VSAs to myriad problems in computational neuroscience, cognitive science and artificial intelligence -- from visual scene analysis to natural language understanding and analogical reasoning.  

\subsection{Implications for neuroscience}

The ability to solve factorization problems is fundamental to both perception and cognition.  In vision for example, the signal measured by a photoreceptor contains a combination of illumination, surface reflectance, surface orientation, and atmospheric properties that essentially need to be ``demultiplied'' by the visual system in order to recover a representation of the underlying causes in a scene \citep{barrow1978recovering,adelson1996perception,barron2014shape}.  The problem of separating form and motion may also be posed as a factorization problem~\citep{cadieu2012learning,memisevic2010learning,anderson2020highacuity}.
In the domain of language, it has been argued that a factorization of sentence structure into ``roles'' and ``fillers'' is required for robust and flexible processing \citep{smolensky1990tensor, jackendoff2002}. Many cognitive tasks, such as analogical reasoning, also require a form of factorization \citep{hummel1997distributed, kanerva1998large, plate2000analogy}.
However, to date it has been unclear how these factorization problems could be represented and solved efficiently by neural circuits in the brain.  VSAs and resonator networks are a potential neural solution to these problems, and indeed developing more neurobiologically plausible models along these lines is a goal of ongoing work. 

In the context of neuroscience and psychology, 
binding is widely theorized to be an important process 
by which the brain properly associates features belonging to the same physical object.
However, how the brain may accomplish this is a hotly debated subject.
Various solutions to this problem, also known as the \emph{Neural Binding Problem}, have been proposed based on attentional mechanisms or neural synchrony 
\citep{treisman1980feature, von1999and, wolfe1999psychophysical}. Note that in these proposals the binding information required to properly describe sets of compound objects has to be added to the individual feature representations, thus increasing the dimension for representing a compound object (or expanding the representation in time). 

VSAs provide a general solution to the binding problem that early visual stages could employ. By using the VSA operations to represent and form data structures, the binding of features is easily expressed. Further, the dimension of the compound representation is not increased. The main computational challenge then becomes the factorization of VSA data structures formed in early sensory pathways, for which the resonator network provides a neurally plausible solution. Interestingly, an earlier discussion of the binding problem by \citet{Feldman2013} has already pointed out that the more fundamental problem of sensory processing is actually one of \textit{unbinding}. Feldman has argued that the raw sensory signals themselves can be thought of as being composites, containing multiple attributes that require factorization, such as in the examples described here. 

In terms of modeling computation in biological neural circuits, resonator networks are clearly an abstraction. In particular, the implementation of VSAs presented here assumes that information is encoded by \emph{dense} bipolar vectors (each element is nonzero), and the binding operation is performed by element-wise multiplication of vectors. 
At first glance these types of representations and operations may not seem very biologically plausible. However, other variants of VSAs that utilize sparse, rather than dense, representations may help to reconcile this disconnect \citep{rachkovskij2001binding,laiho2015high}. 
Recently, we have shown that compound objects can be efficiently represented by sparse vectors with the same dimension as the atomic representations \citep{frady2020sparse}. 
The binding operation in this context relies upon \emph{sigma-pi} type operations \citep{mel1990sigma, plate2000randomly} that are potentially compatible with active non-linearities found in dendritic trees.
Complex-valued variations of VSAs \citep{plate2003holographic} can also be linked to spike-timing codes \citep{Frady2019}, which could further increase links to biology. 

\subsection{Implications for machine learning}

In conventional deep learning approaches, given enough labeled data, a multi-layer network can be trained end-to-end without worrying about understanding or parsing the representations formed by the intermediate layers. Users typically consider the interior of a deep network as a black box. 
However, this conceptual convenience becomes a disadvantage when it comes to improving the deficiencies of deep learning methods, namely susceptibility to adversarial attacks, the need for large amounts of labeled data, and a lack of generalization to novel situations. 
Moreover, while most machine-learning algorithms are focused on problems of pattern matching, or learning a mapping from inputs to outputs, most problems in perception and cognitive reasoning require more than just pattern matching -- they also the ability to form and manipulate data structures. 

VSAs offer a transparent approach to forming distributed representations of potentially complex data structures that may be flexibly recombined to deal with novel situations. For any desired computation, the relevant elements in the data structure can be exposed, or decoded, and combined with other information to calculate a result. 
Here we have shown how these data structures can be formed and manipulated to solve challenging computational problems such as tree search or visual scene analysis.  The key to solving these problems relies upon the ability to factorize high-dimensional vectors, which can now be done by resonator networks.  Given that the problem of factorization arises in many other machine-learning settings, such as simultaneous inference of multiple variables, inverse graphics, and language, it seems likely that resonator networks could provide an efficient solution in these domains as well.

One can potentially combine VSA's with deep learning to get the best of both worlds. An example of this may be seen in our solution to parsing a visual scene (Section \ref{sec:encoding_viz_scenes}). 
Rather than training a network to simply map images to class-labels, our approach trains the network to map the image to a symbolic description that captures the compositional structure of a scene -- i.e., multiple objects combined with their properties -- which can be used by downstream processes to reason about the scene.
Importantly, because multiple object-property bindings can be superposed in the same space, the VSA encoding can handle the very large combinatoric space of possible scenes (in this case, 250 million) with a single vector of fixed dimensionality (500).  
The VSA representation does have a limited capacity and will begin to break down for more than a few objects.  However it is worth noting that human working memory has similar limitations \citep{miller1956magic}.

While there are undoubtedly alternative deep learning approaches for performing analysis of simple scenes, our goal here was to show how analysis of visual scenes could be approached by expressing the problem as a problem of factorization. Incorporating factorization into problems like scene analysis may enable reasoning in much more complex spaces, as such a system can utilize factorization to handle a very large combinatoric space. However, the simple hybrid approach presented here still has some shortcomings, such as requiring a large amount of training data to learn the encoding. 

We believe that multi-layer neural networks could be improved profoundly by enabling all layers to explicitly represent, learn, and factorize data structures. 
Some recent model innovations follow this direction, particularly the  ``Transformer'' neural network architecture which encodes key-value pairs for modeling language and other types of data \citep{vaswani2017attention, devlin2018bert}.
Other model proposals enable the encoding of multiplicative relationships between features using the tensor product \citep{nickel2011three, socher2013reasoning}. 
VSAs could enable these models to represent and manipulate increasingly complex data structures, but this requires solving factorization problems. 
Resonator networks could thus serve as a critical component for building trainable neural networks that form, query, and manipulate large hierarchical data structures.

\subsection*{Acknowledgements}
We would like to thank members of the Redwood Center for Theoretical Neuroscience for helpful discussions, in particular Pentti Kanerva, whose work on Vector Symbolic Architectures originally motivated this project. This work was generously supported by NIH grant 1R01EB026955-01, the NSF grants IIS1718991 and DGE1752814, the Intel Neuromorphic Research Community, Berkeley DeepDrive, the Seminconductor Research Corporation and NSF under E2CDA-NRI, and DARPA's Virtual Intelligence Processing (VIP) program.

\bibliographystyle{apalike}
\bibliography{main}

\end{document}